Review article

# Performance evaluation of deep learning models for image analysis – considerations for visual control and statistical metrics

Christof A. Bertram[1, *], Jonas Ammeling[2, *], Alexander Bartel[3], Gillian Beamer[5], Marc Aubreville[4]

[1] University of Veterinary Medicine Vienna, Vienna, Austria

[2] Technische Hochschule Ingolstadt, Ingolstadt, Germany

[3] Freie Universität Berlin, Germany

[4] Hochschule Flensburg, Flensburg, Germany

[5] Johnson & Johnson Innovative Medicine, San Diego, CA, USA

* Equal contribution

**Corresponding author:**

Christof A. Bertram, University of Veterinary Medicine Vienna, Veterinärplatz 1, 1210 Vienna, Austria. Email: Christof.Bertram@vetmeduni.ac.at

**Abstract**

Deep learning-based automated image analysis (DL-AIA) has been shown to outperform trained pathologists in tasks related to feature quantification. Related to these capacities the use of DL-AIA tools is currently extending from proof-of-principle studies to routine applications such as patient samples (diagnostic pathology), regulatory safety assessment (toxicologic pathology), and recurrent research tasks. To ensure that DL-AIA applications are safe and reliable, it is critical to conduct a thorough and objective generalization performance assessment (i.e., the ability of the algorithm to accurately predict patterns of interest) and possibly evaluate model robustness (i.e., the algorithm's capacity to maintain predictive accuracy on images from different sources). In this article, we review the practices for performance assessment in veterinary pathology publications by which two approaches were identified: 1) Exclusive visual performance control (i.e. eyeballing of algorithmic predictions) plus validation of the models application utilizing secondary performance indices, and 2) Statistical performance control (alongside the other methods), which requires a dataset creation and separation of an hold-out test set prior to model training. This article compares the strengths and weaknesses of statistical and visual performance control methods. Furthermore, we discuss relevant considerations for rigorous statistical performance evaluation including metric selection, test dataset image composition, ground truth label quality, resampling methods such as bootstrapping, statistical comparison of multiple models, and evaluation of model stability. It is our conclusion that visual and statistical evaluation have complementary strength and a combination of both provides the greatest insight into the DL model's performance and sources of error.

# Einleitung

Artificial intelligence, in particular deep learning (DL), holds great promise for revolutionizing the analysis of microscopic images. Indeed several studies suggest that DL-based automated image analysis (DL-AIA) can improve accuracy, reproducibility and/or efficiency of histological evaluations.[7,9,24,62] While there is an increasing body of literature supporting DL-AIA, some veterinary diagnostic and toxicologic laboratories have already incorporated or are considering to implement DL-AIA algorithms into their routine workflows. Similarly, it is expected that researchers will increasingly use these DL tools at a large scale and repeatedly for multiple study populations (not only one-off applications for a standardized study population) to generate biomedical data and gain groundbreaking insights reproducibly across studies. Regardless of these use cases, it is essential to ensure that the quality of DL-AIA is non-inferior to traditional methods, as established through qualified performance evaluation.[10,25,29,38] However, due to a lack of recommendations and guidelines, performance evaluation in veterinary pathology is currently conducted with highly variable rigor. As a result, many pathologists remain skeptical about algorithmic performance, are concerned about potential errors, and are uncertain about the extent to which DL-AIA can be trusted when applied beyond the initial proof-of-principle study. The objective of this commentary is to raise awareness about appropriate methods for performance evaluation and validation before its application and to compare advantages and limitations between visual control and statistical evaluation.

# Why is performance evaluation and validation important?

When DL models are used by pathologists during routine histological examinations, they are diagnostic tests and have to be validated like any other diagnostic test would. To ensure that the initially defined goals of the DL models are met, performance evaluation is important. This typically includes an at least non-inferior or even increased diagnostic accuracy, which can be measured and quantified through multiple parameters. High performance parameters will provide assurance that the training methods and process were appropriate, and that the final model can be used for its intended use case. Since DL model training is an iterative process that uses data to optimize the parameters of a model, the performance of a trained algorithm is highly dependent on many factors such as the training data (including the number of images, image variability, and the quality of the ground truth), the model architecture, and the training process itself (including optimization process, data sampling strategies, and final model selection). Even if all these factors remained the same over multiple training instances, the performance may vary due to the stochastic nature of training (e.g., random initialization of weights, random selection and augmentation of images etc.).

The subsequent article will focus on performance evaluation of the level of the pattern of interest (image classification, object detection, semantic and instance segmentation), whereas we only briefly refer to other evaluation methods. For example, for image classification of tumor malignancy, the question would be if the model had correctly classified all images as benign vs. malignant. Performance of an exemplary object detection model for mitotic figures is evaluated by checking if the location of the predictions match with true mitotic figures.[7] Segmentation can, for example, be used to measure the size of neoplastic nuclei and it needs to be ensured that most of the corresponding pixels are correctly predicted.[24]

The extent by which the DL model performance should be evaluated depends on many aspects such as the complexity of the model and pattern of interest, the degree of intended investigation (e.g., proof-of-concept experiment vs. in-depth scientific study), and prior evaluation of the model (e.g., development from scratch vs. transfer to a new laboratory) and the intended application of the DL model. For example, relatively less rigor may be appropriate for a model developed to count immunohistochemically positive cells for a one-off research application using a specific study set with highly standardized methods of slide preparation. In this case the model would likely be designed to simulate the interpretation (positive vs. negative cell) of a single study pathologists and few uncontrolled sources of error (e.g., differences in digitization devices) would be expected. In contrast, substantially higher rigor of performance evaluation would be required for a mitotic figure object detection model that is routinely used by multiple diagnostic laboratories across a wide range of tumor types. In that scenario, it would be necessary not only to evaluate whether the DL model can handle various preanalytic variables (e.g., different digitization devices), but also to ensure that various pathologists with distinct diagnostic decision thresholds (see below) understand the model's capabilities and how to interpret its predictions. Given this wide range of required rigor for performance evaluation, we do not make specific recommendations in this article as to which method is required as a minimum standard; instead, we explore methods suitable when high requirements for performance evaluation are imposed. These considerations are summarized in Supplemental Table S1.

**Table 1.** List of different methods for performance evaluation of deep learning-based image analysis algorithms.

| Type of testing | Comparison | Type of dataset | Main purpose |
|---|---|---|---|
| **Primary performance indices** | | | |
| Statistical evaluation | Correctness of algorithmic predictions compared to the ground truth using metrics | In-domain test images [a] | Determination of generalization performance to evaluate if the algorithm can predict the pattern of interest sufficiently well |
| | | Out-of-domain test images [b] | Evaluating of the performance robustness towards realistic changes in image features (domain shift) |
| Visual control | Pathologist inspection of predictions on test images (eyeballing) | In-domain test images [a] | Identification of main sources of error |
| | | Out-of-domain test images [b] | Identification of the main sources of error related to domain shift |
| **Secondary performance indices** | | | |
| Measurement accuracy | Comparison / correlation to other tests (such as pathologists' scores) | Application dataset [c] | Evaluation of the consistency to established tests |
| Biological effect or clinical utility (such as outcome analysis or comparison of treatment groups) | Discriminability of case subgroups (defined through biomedical patient characteristics) by algorithmic predictions | Application dataset [c] | Evaluation of the reasonableness of the algorithmic application |

[a] In-domain test images share the same characteristics as the training dataset.
[b] Out-of-domain test images are characterized by distinct features as compared to the test data (such as a different type of whole-slide image scanner).
[c] The application dataset refers to any dataset on which inference of the final algorithm is done to generate analytic data. Even though ground truth annotations are not needed for analysis of secondary performance indices on this dataset type, it may contain the same images as the test dataset but must remain independent of the data used during model development.

There are different methods to evaluate model performance, with each assessing different aspects of the performance (Table 1). One of the main problems is that inappropriate DL techniques can create models which are remembering the data they were trained on or learn the simplest correlation of the training images with the output (spurious correlation; e.g. identify malignant tumors by the presence of ink at surgical margins and benign tumor by the lack thereof), instead of learning the complex decision criteria needed to perform the desired pathology task. Independent datasets used for training and testing are the only way to distinguish these effects. For this reason, it is considered mandatory to evaluate performance on independent hold-out test datasets.[10,29] These independent datasets can be classified into "in-domain" and "out-of-domain" datasets.[10,46]. The in-domain dataset has similar characteristics to the data used for training the model, i.e., it is typically produced alongside the training dataset but separated as a hold-out set before initiating training. Evaluating performance on the in-domain test set allows to establish the generalization performance of the DL model to new cases, i.e., whether the model had learned features that are truly relevant to almost exclusively identify the pattern of interest or whether the features overlap with other patterns (many false positives) or are too narrow (many false negatives).[44] Generalization performance is measured by calculating the number of algorithmic predictions being correct or incorrect, similar to how pathology residents are tested at the end of their training. Testing on images from independent patients (i.e. that were not used for training) will reveal if the model had learned spurious correlations or just remembered the training data. A detailed error analysis helps to identify common failure patterns, where the model had missed to learn discriminatory features, or mistakenly learnt features that are unsuitable for discrimination of the patterns of interest.

The out-of-domain dataset, on the other hand, has distinct features compared to the training dataset, which may cause a so-called "covariant domain shift" in the image representation. Differences between training data and out-of-domain test datasets may be related (to varying relevance) to differences in multiple aspects: animal species, disease conditions/entities, case inclusion/exclusion criteria, sectioning thickness and staining composition (batch effect or inter-laboratory variability), different digitization devices (whole-slide image scanner company and type), digitization settings and image processing.[2,3,55] Image sets created by different laboratories ("external" data) should always be suspicious of having some degree of domain shift, whereas changes within one's own laboratory workflow (such as a different whole-slide image scanner) can also induce a domain shift for "internal" image sets. Testing performance on the out-of-domain dataset evaluates primarily the robustness of the model, i.e., whether the learned features are also relevant to other image domains or are overly specific to the training domain. This information is critical when models are intended for widespread use in multiple laboratories or diagnostic settings. However, even if a broad robustness of the model had been demonstrated, it is recommended that every laboratory should ensure that the DL-AIA algorithm is appropriate for their actual "real-world data".[25]

Besides performance measures on the level of the pattern of interest as described above (e.g., the model detects 9 out of 10 true mitotic figures and detects 2 non-mitotic figures in one tumor image), performance can be indirectly judged on higher levels that reflect the diagnostic application of the model (secondary performance indices). One example would be the measurement level; following the example above, the algorithmic mitotic count is 11, while the pathologist's count is 10 with an absolute error of 1. Errors or consistency with traditional evaluation methods (i.e. pathologists counts) can be further evaluated by Bland-Altman-Plots or correlation analysis. Another secondary indicator would be to evaluate whether the algorithmic mitotic counts can discriminate between patients that die related to the tumor or and those who survive (survival analysis).

Whereas the abovementioned primary and secondary performance measures are often conducted for the "preclinical" evaluation, which is done in conjunction with model development, DL-AIA algorithms need to be further validated when implemented into the laboratory workflow ("clinical validation").[25,46] While these further validation steps are beyond the scope of this article, we want to highlight some additional considerations that are relevant for a reliable use of DL-AIA: performance on real-world data of the specific laboratory, computational efficiency of the model, suitability of user interface and model output (visualization), bias in human-computer-interaction, establishment of standard operating procedures for AIA usage, user training, and ongoing quality management.[25,46,52]

## Why is performance evaluation challenging?

DL-AIA are developed to overcome diagnostic challenges of the current workflow, which includes limited time availability of pathologists for quantitative measurements or shortcomings in diagnostic accuracy (for example by only using a categorical test instead of a numerical test) and low inter-observer reproducibility. For the latter two objectives a superior model performance (in comparison to evaluation of a pathologist) is intended, which can be quite difficult to demonstrate considering the challenges to establish a perfect ground truth that is used as output during model training and as a reference standard for performance evaluation. For most patterns of interest, pathologists are considered the gold standard, but humans are prone to various biases that challenge impartial assessment.[1] Thus, inaccuracy and inconsistency in the ground truth labels can hardly be avoided.[11,23,61] Consequently, most models are developed and tested with an imperfect ground truth while the model developers aim at outperforming this reference standard, resulting in a catch-22 situation (Figure 1). Even if a DL model would perfectly simulate the annotator's decision threshold, the performance evaluation would still indicate some errors, which, in reality, would be random errors in the ground truth of the test set rather than algorithmic errors. Only rarely can a truly superior gold standard be established, such as when histological images are classified based on their molecular signature.[49]

Another highly relevant challenge is the marked differences in pathologists' interpretations, particularly regarding the setting of thresholds between label classes or between patterns of interest vs. imposters.[7,9,24,61,62] An algorithm is likely to reflect all biases of the training data, including the individual decision threshold applied by the dataset annotator. However, if the algorithm is used by another pathologist with a differing individual decision threshold, an unjust perception of poor algorithmic performance may arise, which, however, mostly reflects the mismatch between the two pathologists (i.e., the model user and dataset annotator) and not necessarily the incapability of the DL model. In our experience, pathologists often expect perfection from DL models while remaining unaware of their own inter-rater variability which may not have been accounted for in the training dataset's ground truth.

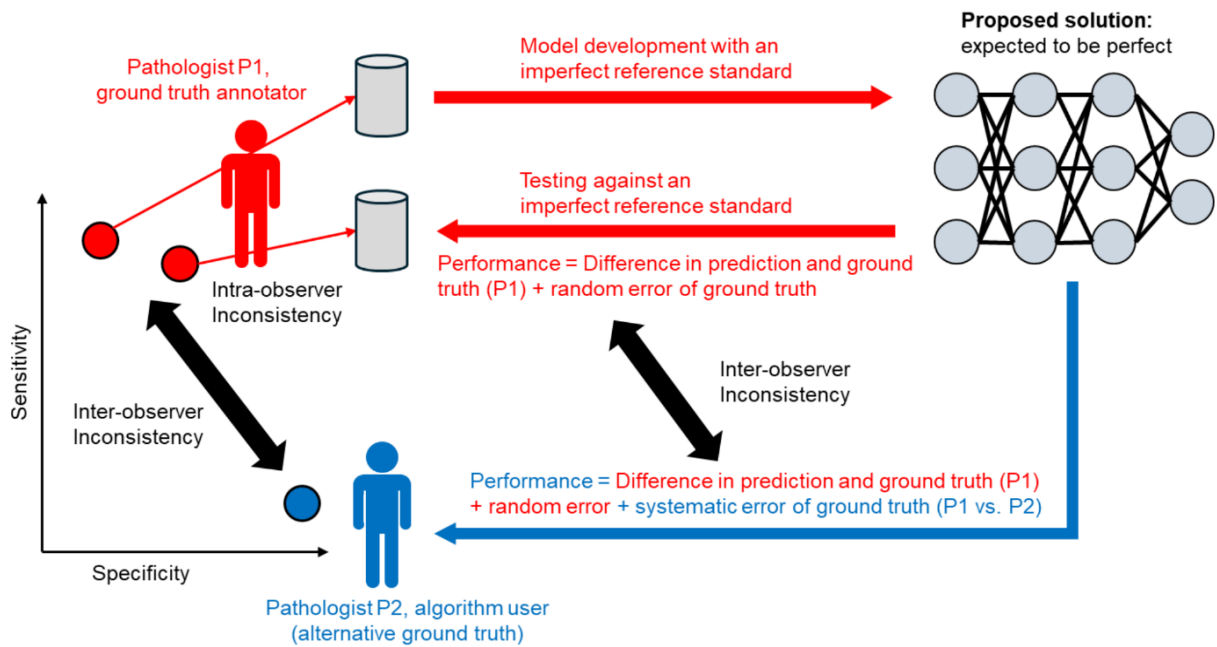

**Figure 1.** Catch-22 situation of deep learning-based model development and testing. One of the diagnostic problems that a deep learning-based model may be expected to overcome is the inter- and intra-observer inconsistency of pathologists. For example, one pathologist may have higher sensitivity but lower specificity with some intra-rater disagreement (two red dots), while another pathologist may have a lower sensitivity, but higher specificity (blue dot). Despite this well-known difference in pathologists' interpretations, they are paradoxically used to establish the reference standard (ground truth). As the model is trained with imperfect data and, more importantly, is evaluated against an imperfect reference, it becomes challenging to demonstrate the superiority of the model as compared to the pathologist. Based on the red pathologist's decision thresholds, the resulting model is expected to be rather sensitive, whereas 100% performance of the model cannot be tested due to random errors in the round truth. To complicate matters further, a second pathologist (blue, in this case the more specific rater), may criticize the algorithmic performance due to a perceived low specificity of the model. However, this difference in the performance reception is primarily related to the differences between the two raters - precisely the issue the model was developed to address in the first place.

To support our claims about intra- and inter-rater inconsistency, we draw on published literature. These findings intend to show that some microscopic patterns are extremely difficult due to substantial morphological overlap with look-alikes which is coupled with different thresholds between observers in discriminating between pattern of object and imposter. While numerous studies have demonstrated the challenges of reproducibility in quantitative pathology tasks, the three examples below also illustrate how annotation variability can influence the performance evaluation of DL models. In the first study, object detection models for mitotic figures were evaluated using the open access dataset from the TUPAC challenge with annotations by the initial dataset creators and an alternative ground truth by a second research team.[11] A comparison of the two dataset revealed that both included labels for the same 1,239 objects, while 313 and 760 objects, respectively, were unique to one dataset. When the same DL model development approach was applied, the measured performance differed from F1 = 0.549 to 0.735 (with 1.0 indicating perfect performance), solely depending on which ground truth definition was used. The second study applied a similar mitotic figure detection model, which had been shown to have a high performance on the study population (immunohistochemistry-guided ground truth) with an F1-score of 0.83.[7] Twenty-three study pathologists evaluated the same images and counted between 1263 and 4412 mitotic figures, i.e. they differed by up to a factor of 3.5. Their performance in relation to the ground truth ranged between F1 = 0.53 and 0.79, with some being highly consistent with the DL model and others having markedly differing decision thresholds (more sensitive or more specific). These pathologists with distinct decision thresholds are likely to be sceptical about algorithmic performance, even though the model is more consistent with the ground truth than their interpretation threshold.

The third study developed an object detection model for binucleated cells in mast cell tumors.[8] To establish a benchmark for algorithmic performance, six pathologists annotated binucleated cells in a small subset of the test set. Compared to the ground truth, which was established by a different annotator (with a total of 148 ground truth annotations), the six pathologists additionally annotated ("false positives") between 56 to 542 objects. Depending on which "truth" was used for testing the model (majority vote of benchmark pathologists or the initial single annotator ground truth), the measured performance of the same model varied between an F1-score of 0.424 (majority vote) to 0.667 (initial ground truth). On the initial ground truth, the benchmark pathologists archived a F1-score between 0.270 to 0.526. If the pathologists with the lower metric values would evaluate the model's performance it would be their impression that the model had overlooked numerous binucleated cells, whereas this biased interpretation would be related to the lower decision threshold (i.e. higher sensitivity) of these pathologists as compared to the ground truth and DL model.

From a statistical perspective, the limitations of the reference standard can be conceptualized as two forms of uncertainty: aleatoric and epistemic.[31] Aleatoric uncertainty reflects inherent ambiguity in the data, such as genuine morphological overlap between patterns of interest and imposters, where even highly experienced pathologists may reasonably disagree. Epistemic uncertainty, in contrast, arises from incomplete knowledge or suboptimal processes, such as poorly defined diagnostic criteria, inadequate training, or inconsistent annotation protocols. Both uncertainty types are embedded in the ground truth used for DL model development and testing, and they jointly determine the "noise level" of the labels. When this label noise is high, even a conceptually perfect model cannot reach an apparent performance of 100% relative to the reference standard, because some fraction of apparent "errors" are in fact disagreements with inconsistent or ambiguous labels rather than genuine algorithmic failures.

Quantifying this label uncertainty is important for placing DL-AIA performance into context. Such a benchmark can be either defined by comparison by the consistency of the annotation workflow or by multi-rater studies. Consistency of the test set ground truth can be determined if the same annotation workflow is applied a second time to the test set images. The difference between the labels of the two datasets will be a good indication of random error / aleatoric uncertainty and thereby approximate an upper bound on achievable algorithmic performance. For example, if the agreement on the object-level between the two time points is F1 = 0.8, then it is unreasonable to expect a DL model applied to the same labels to substantially exceed this agreement. In such settings, a model achieving an F1-score close to the level of agreement may already represent near-optimal performance, even though the metric itself appears far from perfect. Multi-rater studies, in which several pathologists independently annotate the images of the test set and are compared to the ground truth, can be used to estimate the inter-rater reproducibility of the reference standard and thereby approximate the systematic error / epistemic uncertainty. Inter-rater agreement statistics (e.g., pairwise F1-scores, Kappa coefficients, or majority-vote comparisons) may provide an estimate of the lower performance boundary for the model to be non-inferior to pathologists.

These considerations emphasize that performance metrics must always be interpreted relative to the quality and consistency of the ground truth. When reading DL-AIA studies, it is therefore essential to consider how the reference standard was created (e.g., single annotator vs. consensus or majority vote, use of auxiliary stains for decision support, computer-assisted candidate detection) [12] and whether any estimates of inter- or intra-observer variability are available. Ideally, studies should report both human-ground truth and model-ground truth agreement on the same test images, allowing the reader to judge whether the algorithm has reached, matched, or exceeded typical expert consistency for the task at hand. Without this benchmark, a performance metric such as F1-score of 0.7 remains difficult to interpret: it may indicate poor performance on a simple, highly reproducible task, or conversely, very strong performance on a task for which human agreement is intrinsically limited by high aleatoric uncertainty.

## Current practice of performance evaluation

The performance of the DL model on the pattern level can be evaluated through statistical methods and visual inspection (i.e., eyeballing of the algorithmic predictions by a pathologist).[10] Statistical performance evaluation involves comparing ground truth annotations (the reference standard) with algorithmic predictions on the test dataset, followed by the calculation of performance metrics.[25,38] This article does not discuss the various metrics, as a dedicated publication on this topic already exists.[38] Visual evaluation, on the other hand, is performed by a pathologist who reviews and interprets algorithmic predictions on the test dataset, estimating the proportion of correct and incorrect predictions and identifying the primary sources of error. Currently, there are no established guidelines in veterinary pathology regarding when and how to apply these methods. Therefore, this section analyzes current practices in research articles published in selected veterinary journals.

A literature search was conducted on the websites of 5 journals focusing on veterinary pathology using the following search terms: "artificial intelligence", "machine learning", and "deep learning". Original articles published between 1.1.2015 and 3.9.2025 that describe the development of DL-based models for microscopic images (histology and cytology, histochemical stain and immunohistochemistry) and/or the evaluation of such a DL model's performance on any of the levels listed in table 1 were included. We identified 25 articles in *Toxicologic Pathology*,[4,6,16-18,22,26,27,30,32-34,36,41-43,45,48,50,51,53,54,58,59,63] 6 articles in *Veterinary Pathology*,[7,9,21,24,49,62] and no eligible articles in the *Journal of Comparative Pathology*, *Veterinary Clinical Pathology*, and *Journal of Veterinary Diagnostic Investigation*. From these articles we have extracted information on 1) the type of AI utilization, as defined for the *Veterinary Pathology* reporting guideline,[10] 2) the use of different dataset types for performance evaluation, as defined by a recent review article,[12] 3) the independence of the datasets used for performance evaluation from the development dataset, and 4) if visual and statistical performance evaluation was conducted.

Of the 31 eligible articles, most (N = 30) had developed new DL models for their studies, either for a previously unexplored use case (Type 5 of AI utilization; N = 28) or by using an established development process for this use case (Type 4 of AI utilization; N = 2). These 30 studies used a primary dataset with ground truth annotations, for development of the DL model (N = 30) and testing its performance (N = 17, 57%). Additionally, performance evaluation was done on a secondary test set (dataset with ground truth annotations created independently from the primary dataset) in 4 instances and an application dataset (dataset without ground truth annotations, but with biomedical metadata of patients) in 15 instances. Only 1 study validated a previously published DL model on a secondary test set; this study will not be further discussed here. The test datasets of 4/30 (13%) studies are publicly available.

For the 19 studies that used a primary or secondary test dataset with ground truth annotations, the test set was considered independent in 12 instances (63%), while concerns for data leakage (i.e., use of image parts from the same patient in both the development and test subsets) was raised for 7 articles (37%). Visual performance control was reported in 14/19 studies and statistical performance metrics could be found in 18/19 studies, with a benchmark of pathologists' performance (as compared to the ground truth) being available in 4 studies. Secondary performance indices (determined on any dataset type) were additionally provided in 12 studies.

Studies with Type 4-5 DL tools that did not utilize a primary or secondary test set (N = 11/30), restricted their performance evaluation on application datasets with visual assessment (N = 11) and secondary performance indices (N = 11). For these application datasets, concerns of data leakage were found in two studies, whereas three studies lacked information to judge data independence. Terminology used to describe the pathologists' interpretation of visual assessment included "satisfactory", "adequate", "suitable", "frequently failed", and "occasionally missing". This evaluation approach was done in 11/25 (44%) of the articles published in *Toxicologic Pathology*, which aligns with a survey of toxicologic pathologists, which revealed that 40% of respondents exclusively relied, when developing DL models, on visual performance evaluation by a single pathologist.[47]

The dataset creation methods for Type 4-5 studies (reported in 24/30 studies), followed two broad approaches: 1) a priori to model training and 2) stepwise during model training (active learning-like). An a priori dataset creation had almost always (N = 15/16, 94%) resulted in a separate hold-out test dataset with ground truth annotations and 13/15 studies leveraged this test set to calculate statistical performance evaluation. Stepwise dataset creation was done by 8 studies and is motivated by high time-efficiency since the minimum number of annotations, needed for training sufficient performant algorithms, are created. This approach is characterized by cycles of annotation creation, model training, preliminary model inference and visual quality control. In each cycle the training data is augmented by new images and annotations (computer-assisted annotation method) with subsequent continuation of training the preliminary models. The preliminary models' predictions are used in guiding the developer to select particularly challenging images (active learning-like method) and are often used to speed up labeling (computer-assisted annotation method). Whereas ground truth annotations are always needed for model training, 5/8 studies (63%) had not reported establishment of a test set, consistent with a 2021 survey of toxicologic pathologists that found 21% of respondents did not perform data splitting into training, validation, and test sets.[47]. Possibly related to the time-efficient concept, this active learning-like workflow seems to be often associated with avoiding creation of laborious annotations of a test set. When establishing a test set, it is important that the dataset split is done from the first manual dataset (before starting stepwise model creation). If the test set was created through this active learning-like process, the model would be overfit to the test dataset, hindering reliable performance evaluation.

# Comparison of statistical and visual performance evaluation

In veterinary pathology, there is currently no consistent approach to evaluate the performance of DL models. Some authors rely exclusively on visual assessment, presumably to eliminate the time investment required for test dataset creation, while others provide statistical analyses (see above). In human medicine, clear recommendations exist from the College of American Pathologists, which mandate rigorous statistical performance evaluation for all DL-AIA algorithms before their implementation as diagnostic tools for patient samples.[25] Similarly, the *Veterinary Pathology* reporting guideline for research studies on AI-based AIA recommend the use of statistical evaluation in publications unless authors can provide a rational explanation for why metrics are not applicable.[10] This reporting guideline emphasizes that the advantages of statistical performance evaluation will, in most research projects, outweigh the time investment required for dataset creation. However, detailed recommendations for ensuring sufficient performance of diagnostic DL-AIA algorithms for the various veterinary applications are currently lacking. We encourage the veterinary pathology community to engage in further discussions on this topic. To raise awareness and establish a foundation for good scientific practices in our field, this section compares the advantages and disadvantages of visual and statistical performance evaluation approaches (Table 2).

**Table 2.** Comparison of the two performance evaluation methods: statistical assessment (metrics) and visual control.

| Aspects | Statistical assessment | Visual control |
|---|---|---|
| Precondition | Ground truth annotation of the test set with "real-world" variability and high label quality | Trained expert with high pathology expertise; visualization of algorithmic predictions on the image |
| Transparency (how well can the evaluation be communicated?) | High (test dataset and formula of metrics can be clearly defined) | Low (equivocal and subjective terminology such as "sufficient" performance") |
| Interpretability (how well can it be comprehended if the performance is good or poor?) | High (metrics value within a predefined range/scale; benchmark can be established, e.g., defined as the performance of pathologists or previous models) | Low (subjective terminology, a benchmark for the specific dataset cannot be communicated) |
| Reproducibility (can the same evaluation be done multiple times in the same experiment?) | High (automated calculation; consistent dataset and ground truth) | Moderate (possibly different image regions are evaluated, inter-rater variability) → comparison |

| Aspects | Statistical assessment | Visual control |
| --- | --- | --- |
| | → allows a reliable comparison of different models | of different models is unreliable |
| Reproducibility (can someone else conduct the same evaluation?) | High (clearly defined formula, test dataset may be made publicly available) | Low (subjective decision criteria, high inter-rater inconsistency) |
| Accuracy and reliability (can fine differences in model performance be detected?) | High (Fine differences can be detected, e.g., F1-score of 0.801 vs. 0.802; performance can be calculated over a large image set) | Low (lack of nuanced terminology and cognitive limitations to estimate performance over a large image set) |
| Bias (is there a risk that the measured performance markedly differs from the truth?) | Lower (Bias of ground truth can be mitigated, e.g., through blinded multi-expert majority voting) | Higher (human are prone to numerous biases, e.g., evaluation may be influenced by the general attitude towards AI) |
| Ability to compare image subgroups (for example, can a difference of the performance based on two WSI scanner types be evaluated?) | Easy (e.g., F1-score for scanner 1 is 0.801 and 0.652 for scanner 2) | Difficult (subjective interpretation, extend of pairwise comparison is limited) |
| Ability to identify the sources of error (for example, are false detections of mitotic figures related to necrotic areas or inflammatory cells?) | Possible through analysis of subgroups of images or image regions | Possible by classifying the morphology that causes false predictions |
| Time investment (how much time investment is needed to conduct this analysis | High for dataset creation; low for metrics calculation → as many metrics and subevaluations as needed can be calculated with little additional time investment (scalability) | Low, but only for one-time testing of a small image set |

While visual performance evaluation is only time-efficient when done non-exhaustively on a small image set for a few times, it has several limitations compared to statistical methods, primarily due to the subjectivity inherent in expert assessments. Communicating the results of visual evaluation is often restricted to subjective terminology, such as "sufficient" or "suboptimal." In contrast, statistical evaluation requires a ground truth dataset, which is time-consuming to create, but it enables extensive and repeated analyses with minimal additional time investment (scalability). Statistical evaluation produces specific performance metrics, such as an F1-score (e.g., F1 = 0.801), based on clearly defined formulas. Thus, these metrics are unambiguous and can be easily reproduced by other algorithm developers.

The main advantage of visual performance evaluation is its ability to identify the sources of algorithmic error by inspection of the visualization of algorithmic predictions on the images, which is possible for object detection and segmentation tasks, but not for image classification or regression tasks. A trained pathologist can detect shared features of false predictions and rank error sources accordingly. For example, a pathologist might observe that a mitotic figure detection algorithm produces many false positives in inflamed or necrotic areas, suggesting that the training dataset should be augmented with these regions or that such regions should be excluded for model application. However, statistical metrics can complement this process by narrowing down the images worth investigating visually, as they can display results on a per-image basis. Statistical evaluation also has the potential to detect "hidden stratification" within test datasets.[29] For instance, a test dataset comprising images from multiple whole-slide image (WSI) scanners may reveal, through stratified analysis, that images from a particular scanner performs poorly, making it unsuitable for use with this particular algorithm.[2,3]

Both visual assessment and ground truth annotations are prone to human bias, but the extent of bias can be minimized in ground truth annotations by using sophisticated methods, such as blinded majority voting by multiple annotators.[61] In contrast, visual assessment is more susceptible to bias, which can significantly influence performance evaluation for complex patterns of interest. For example, AI enthusiasts may unconsciously overestimate performance by ignoring errors, while AI skeptics may underestimate performance by focusing excessively on errors. Additionally, the threshold for acceptable performance may decline over time as developers invest significant effort into model development, even if performance improvements are minimal. Statistical evaluation, on the other hand, is entirely reproducible when using the same test dataset and ground truth annotations. While statistical evaluation can be applied to large datasets (encompassing a real-world variability), visual assessment is often restricted to selected image regions. This limitation introduces the risk of evaluating non-representative regions. Even if large image regions or multiple images are assessed visually, it is cognitively challenging to estimate average performance across a vast number of decisions.

A significant limitation of visual assessment is its inaccuracy when comparing different models or training approaches. The low number of image regions that can be compared pairwise and the cognitive limitations of humans, such as estimating the proportion of over-segmented tissue, make such comparisons unreliable.[1] In contrast, performance metrics provide high accuracy in detecting fine differences between models. However, the choice of a specific metric can influence the ranking of models. To address this, the use of multiple complementary metrics is generally recommended.[37,38] The choice of appropriate performance metrics depends on several factors: (1) the pattern recognition task (e.g., image classification, object detection, semantic segmentation, or instance segmentation), (2) domain-specific properties (e.g., whether the shape or center of an object is relevant), (3) target structure properties (e.g., whether objects are of similar or variable size), (4) dataset properties (e.g., equal class distribution or class imbalance), and (5) algorithm output-related properties.[38] A decision guide, along with detailed information about the proposed metrics and their limitations, is available on the *Metrics Reloaded* website: https://metrics-reloaded.dkfz.de/. In addition to the pattern recognition task, the pathology task should also be considered when selecting metrics. For example, counting tumor cells in an image can be achieved through object detection or by segmenting nuclei and subsequently counting them. If segmentation is chosen as the pattern recognition task (with metrics such as the Dice Similarity Coefficient), it may still be useful to evaluate object detection performance (e.g., using recall and precision) to ensure alignment with the pathology task.

To properly interpret the performance of a DL-AIA algorithm, a benchmark established for the specific test images is essential. However, benchmarks can only be accurately defined using metrics (i.e. when compared against the ground truth of the test set), not subjective terminology derived from visual assessment. Without a benchmark, it is difficult to determine whether a performance metric, such as an F1-score of 0.7, is good or bad, as this depends on factors such as dataset composition (e.g., the proportion of difficult cases and image quality), task complexity, and ground truth consistency. While perfection of an DL model is not expected, the goal should be to demonstrate improvement over previous tests. This improvement can be defined in two ways: 1) by using traditional tests (e.g., assessments by multiple pathologists) to set a minimum performance threshold [7,9,21,24,62] or 2) by outperforming a state-of-the-art model through optimization of the model development process.[2,3] When test datasets and their ground truth annotations are publicly available, the superiority of a DL model can even be compared across multiple studies. However, care must be taken when applying a model to an external public dataset (i.e. created by a different group), as performance differences may arise not only from a domain shift but also from differences in ground truth definitions.[3,11,61].

## Requirements of the test dataset

The precondition for performance evaluation is the availability of a dataset subset reserved for testing (hold-out test set). Due to the delicacy of performance evaluation, creation of the test set has particularly high requirements for both image quantity and label quality, which are discussed in this chapter.

First, it is critical that the test set is not used for training; images or image parts from the same patient cannot overlap between the different data subsets.[10,29] To ensure that the features learned by a DL model are generally useful for predicting the pattern of interest, generalization performance must be evaluated on a previously unseen hold-out test dataset (i.e., images from the same patient must not be distributed across training and test subsets).[10,29] During training, a model may overfit to the training data, possibly learning highly specific features of those images. If such an overfitted model is tested on other parts of the same images used for training (data leakage), it may exhibit seemingly high performance.[15] However, this performance is misleading, as it is driven by spurious correlations in the test images and cannot be reproduced on an independent test set (i.e. later model application). Concerns about data leakage were identified in some of the evaluated articles (see above). However, even with a dataset split on the patient level, data dependencies cannot be fully excluded (e.g., all tumor slides have inked margins while slides with non-neoplastic lesions lack ink). Risk for data bias can be reduced when training images and testing images were created in separate batches (possibly use a second test dataset created by the same group at a later time point when a more thorough evaluation is required) or at different laboratories (possibly use external images). Consequently, it is advisable to use multiple test datasets (primary and secondary) before routine usage of the DL model.

Careful consideration should be given to the size of the test dataset, as this directly affects the precision and reliability of the performance estimates. In many DL-AIA studies, test set size is driven primarily by data availability or annotation capacity. [29] However, even simple sample size considerations can help to avoid overinterpretation of unstable metrics. For example, a mitotic figure detection algorithm with a sensitivity of 90% based on a small number of cases may appear excellent, yet the corresponding confidence interval could be so wide that substantially lower "true" performance cannot be excluded. In contrast, a similar point estimate derived from a sufficiently large number of independent cases will have a much narrower confidence interval and therefore provide stronger evidence that the performance is acceptable for the intended diagnostic application. From a practical perspective, test set size planning can be framed in terms of the desired precision of key metrics, rather than formal hypothesis testing or statistical power calculations. For binary outcomes (e.g. correct vs. incorrect classification at the case level) approximate binomial methods can be used to estimate how many positive cases are required to achieve a given confidence interval width around sensitivity or specificity. For instance, if a model is expected to have sensitivity of approximately 90% for detecting a particular lesion, and the goal is to estimate with a 95% confidence interval no wider than ±0.05, this implies the need for roughly on the order of one to a few hundred positive cases in the test set, depending on the exact assumptions and method used. Importantly, such calculations should be performed at the level of independent units relevant to the application (e.g., animals, patients, or whole-slide images), rather than at the level of tiles or individual image patches, to avoid artificially inflating the effective sample size. Even though sample size calculation for the test images may be feasible in some instances, it is acknowledged that the size of the dataset is often limited by the time required for its creation, making formal sample size targets difficult to achieve.[25,29] Nonetheless, researchers should explicitly acknowledge these constraints and interpret performance metrics with appropriate caution when test sets are small or unbalanced.

When the default three-way dataset split results in a small, non-representative test subset, the k-fold cross-validation approach may be used as an alternative. In this method, the dataset is divided into several (number indicated by k) subsets (folds), and each fold is used independently for training, validation, or testing across multiple iterations of model development.[46] This approach enables that every image in the dataset is used for testing in at least one fold, thereby increasing the variability of images tested across all folds combined. However, due to the increased computational resources required to develop multiple models, cross-validation is typically reserved for small datasets.

In addition to the sample size, test datasets must also be representative and cover a realistic variance of images. This means that the image features in the test set should reflect those of the real-world use case for which the model is intended.[25,29] For example, if a mitotic figure detection algorithm is intended to analyze entire whole-slide images, it must be tested on images from all regions of a whole-slide image and not only perfect regions of interest (intra-slide variability). At the same time the use of a single whole-slide image with thousands of annotations may be less informative than using multiple random regions of interest from numerous tumor cases, which better encompass the inter-slide variability. Another example is if the model is intended for use in different laboratories, it must be tested on samples prepared in various laboratories and digitized using different whole-slide image scanners. To sufficiently encompass the intended use case, a large and diverse test dataset may be required that encompasses a realistic variability in tissue quality, tissue morphology (e.g., different disease subtypes), slide preparation and digitization methods (e.g. staining, ink, color representation) and artifacts (e.g., tissue folds, scan artifacts). In some cases, oversampling of certain case subgroups (as compared to their actual frequency) may be favorable to include all relevant morphological subtypes and allow for stratified performance evaluation of relevant variables.[35] In contrast, active-learning based image selection and synthetic images should not be used for test datasets, since they do not ensure a realistic and real-world image representation.[12]

Previous analysis of AI articles has highlighted the importance of test set size and variability, showing that removing a single case from a small test set can significantly alter performance metrics.[37] Statistical methods such as bootstrapping and confidence intervals for performance metrics may help mitigate this effect. The analytical derivation of confidence intervals can be challenging, particularly for complex metrics or structured data. Therefore, resampling methods such as bootstrapping are highly attractive, as they can be applied to a broad range of metrics with relatively few assumptions. In a case-level bootstrap, the test dataset is repeatedly resampled with replacement at the level of independent units, and the metric of interest is recalculated for each resample. The empirical distribution of these bootstrapped metric values can then be used to derive confidence intervals.[20] Resampling approaches are also useful when comparing different DL models or training strategies on the same test set. In such paired settings, bootstrap methods can estimate the distribution of the difference between two models' performance metrics across resamples, allowing the calculation of confidence intervals for these differences. This enables an assessment of whether an observed performance improvement is likely to be meaningful or falls within the range of random variation due to the finite and possibly heterogeneous test dataset.

The quality of the ground truth is a critical consideration for test datasets. As discussed above, ground truth annotations by pathologists are inherently subject to bias, and errors in the ground truth of the test set can unjustly diminish algorithmic performance. A well-planned annotation workflow (including detailed annotation instructions with clear definitions of the label classes) and, depending on the complexity of the pattern of interest, advanced annotation methods may be necessary to reduce annotation biases.[12] These methods may include majority voting by multiple annotators,[61] computer-assisted identification of missed candidates,[11] or decision guide by providing additional staining through image registration.[23] On the other hand, algorithmically generated labels (to speed up the annotation process) may diminish label quality and careful review by experts is critical.[39]

Finally, authors of articles on DL-AIA should consider making their test datasets (or entire datasets) publicly available. Open access to these datasets allows other researchers to evaluate their algorithms on external images, facilitating comparisons and providing a valuable resource for assessing algorithmic robustness.[35] However, systematic differences in the ground truth between the in-house and external dataset must be considered as potential sources for a drop in performance.[11] Considering the paucity of open datasets,[12] a relevant limitation of current DL research is that only a few datasets are frequently used for the development of state-of-the-art DL methods.[57] While a continuous, slight improvement in performance may be reported in the literature over time, this may reflect hyperparameter overfitting to a specific dataset rather than an actual improvement in the DL method. This consideration highlights the need for more researchers to make their dataset accessible to facilitate performance evaluation of innovative DL-AIA methods.

# Statistical comparison of deep learning models on the same test dataset

In many DL-AIA studies, multiple model development methods are compared to identify the approach that yields the best performance, which is ideally supported by statistical comparisons. The differences in the model development may be related to different model architectures, alternative training strategies, varied hyperparameters, or the inclusion of domain-specific augmentation strategies. These models are usually tested on the same hold-out test set. This means that for each animal, patient, or whole-slide image in the test set, every model makes a prediction on exactly the same case. The results of two models are therefore not independent, they are paired case by case. This paired structure is important, because it allows more sensitive and appropriate statistical comparisons that directly examine whether the performance of two models is actually different. These considerations, and suitable statistical methods for exploiting such paired structures, are described in the following paragraphs.[19]

If performance metrics are computed at the level of individual cases (e.g., per-patient accuracy, per-slide Dice coefficient), then paired statistical tests are appropriate. For classification tasks at the case level (e.g. correct vs. incorrect diagnosis per animal or whole-slide image), McNemar's test is a widely used paired method to assess whether two models differ significantly in their error rates.[19,40] McNemar's test evaluates the discordant pairs, where one model is correct and the other is incorrect, and detects systematic differences in performance even when overall accuracy appears similar. For tasks that produce continuous or count-based outputs at the case-level, such as mitotic counts, cell densities, or segmentation overlap scores, paired statistical tests such as the paired t-test or the Wilcoxon signed-rank test can be used to compare models. These tests assess whether the distribution of per-case errors or measurements differs systematically between two models.

When multiple cases, classes, or regions of interest are involved, performance can be summarized using either macro- or micro-averaging, and the choice should be guided by the clinical relevance of the diagnostic task.[7] Micro-averaging aggregates all individual predictions across the dataset (e.g. all candidate mitotic figures) before computing a single global metric (e.g. one overall F1-score), so that each prediction contributes equally to the final score. In contrast, macro-averaging first computes a performance metric for each independent unit of analysis (e.g., per slide, per patient, per tumor subtype) and then averages these unit-level metrics, giving each unit equal weight regardless of how many predictions it contains. For the task of mitotic figure detection, micro-averaging is often used because it reflects overall object-level detection performance, where each candidate mitotic figure is treated equally.[2]

However, macro-averaging could better capture clinically relevant unit-level effects, for example by penalizing ten false positives on a slide with no mitotic figures more strongly than the same number of false positives on a slide with 100 mitotic figures. If unit-level (macro-averaged) metrics are considered most relevant, the resulting per-unit scores naturally lend themselves to paired statistical tests such as the paired t-test, or the Wilcoxon signed-rank test. If only a single micro-averaged aggregate metric per model is of interest, the paired structure at the unit level is no longer directly available, and comparisons should instead rely heavily on the confidence intervals of the aggregate metrics for each model. If the confidence intervals for two models' aggregate metrics (e.g., 95% CI for Model A's F1-score vs. 95% CI for Model B's F1-score) show substantial overlap, it suggests that there is no statistically significant difference in their performance. Conversely, clearly separated confidence intervals would indicate a statistically discernible difference. These confidence intervals for aggregate metrics can be derived using resampling methods like bootstrapping, where the entire test dataset is resampled with replacement multiple times as described above.[20]

Regardless of whether case-level or aggregate metrics are used, it is recommended to complement hypothesis tests with effect size estimates and confidence estimates for the difference in performance metrics (e.g., Δaccuracy, ΔF1-score, or ΔAUC). These can be obtained through case-level bootstrapping (for case-level metrics) or by comparing the distributions of bootstrapped aggregate metrics (for global metrics), yielding an empirical distribution of the performance difference. This provides more interpretable information about whether one model offers a meaningful improvement over another.[56]

A common pitfall in model comparison arises when many models or configurations are tested on the same dataset, leading to a multiple comparison problem. For example, if 20 different hyperparameter configurations are each compared to a baseline model using a significance threshold of p < 0.05, one would expect at least one "significant" result by chance alone, even if no true differences exist.[5] This inflates the risk of falsely concluding that a particular model is superior. When multiple pairwise comparisons are conducted, appropriate corrections (e.g. Bonferroni, Holm, or false discovery rate adjustments) should be applied, or the analysis should be reframed as an exploratory investigation rather than confirmatory hypothesis testing.[5,28] Alternatively, researchers may choose to prespecify a small number of key comparisons or rely primarily on effect sizes and confidence intervals, which are less prone to inflation from multiple testing and provide more interpretable evidence of practical differences.[56,60]

An additional caveat applies when test datasets are very large or when performance is evaluated at the level of thousands of tiles or individual objects rather than independent cases. In such scenarios, even trivial differences in performance can yield highly significant p-values, leading to overinterpretation of negligible improvements. For this reason, statistical comparison should always be anchored in clinical or biological relevance (e.g., whether the observed difference in performance translates into meaningful change in diagnostic accuracy, workload reduction, or patient outcome). Visual inspection of error distributions, stratified analyses by case difficulty or image domain, and secondary performance indices, such as correlation with pathologist scores or discriminability of prognostic groups, can all help contextualize whether a statistically significant difference is also practically important.

# Multiple training runs and statistical comparison of model stability

Due to the stochastic nature of DL model training (e.g., random weight initialization, stochastic optimization, random data augmentation), repeated training of the same architecture on the same data can yield slightly different models with different performance metrics. Consequently, a single training run may not adequately reflect the typical performance of a particular model configuration. To better characterize model stability and to avoid overinterpreting potentially optimistic or pessimistic single-run results, it is advisable to train each candidate model configuration multiple times with different random seeds and to summarize performance across these runs.[13,14]

In many practical DL-AIA applications, however, computational constraints limit the number of independent runs to a relatively small number (e.g., 3-5 per configuration). When only a single aggregate performance metric (such as global F1-score or global Dice coefficient over the entire test set) is available for each run, the resulting sample size for between-configuration comparison is therefore very small. While, in principle, paired statistical tests (such as a paired t-test or Wilcoxon signed-rank test) could be applied to the per-run metrics of two configurations, the low number of runs leads to limited statistical power and unstable p-values, making such hypothesis tests difficult to interpret robustly. In this context, repeated runs are best regarded primarily as a model stability and sensitivity analysis, demonstrating whether performance is broadly consistent across seeds, rather than as a basis for strong inferential claims about the superiority of one configuration over another. Large variability across runs suggests that the model is sensitive to random factors and may not be reliable in routine use, even if the best-performing run shows high metrics.[14]

More formal statistical comparison of model configurations benefits from having either a larger number of independent runs per configuration or the ability to resample the test data at the level of independent units (e.g., animals, patients, whole-slide images). In the latter case, bootstrap resampling of the test set can be combined with multiple training runs to account for both sources of variability: for each of K training runs, the test set is resampled with replacement B times (e.g., B = 1000) at the case level, and the aggregate performance metric is recalculated for each bootstrap sample. This yields a distribution of performance metrics that reflects both training stochasticity (across runs) and sampling variability (across bootstrap resamples). Confidence intervals for performance metrics and for differences between configurations (Δmetrics) can then be derived from these combined distributions, providing a more nuanced and reliable assessment than single p-values.[19,20] When sufficient K × B samples are available, formal comparisons can be based on the empirical distribution of Δmetrics (e.g., bootstrap confidence intervals or non-parametric paired tests), but interpretation should still focus on effect sizes and confidence intervals rather than p-values alone, particularly when many model configurations or hyperparameter settings are explored in parallel.[56,60] In all scenarios, emphasis should be placed on the magnitude and clinical relevance of observed differences, with repeated runs serving to demonstrate that reported performance is stable rather than the result of a single favorable training outcome.

## Conclusion

In conclusion, proper performance evaluation of DL-AIA algorithms is essential before their implementation into research or diagnostic workflows to ensure safe and effective use. Numerous challenges exist, particularly because these models are typically (with few exceptions) compared to pathologists' decisions and therefore inherit all limitations of an imperfect reference standard, including inter- and intra-observer variability and ambiguities in diagnostic thresholds. This comparison can be conducted either through visual assessment or by calculating the differences between predictions and ground truth annotations (statistical evaluation), and both approaches are currently used in veterinary DL-AIA publications.

This article highlights that exclusive reliance on visual assessment is not sufficient for evaluating complex DL models and has substantial limitations in terms of transparency, reproducibility, and the ability to communicate performance in a manner that is comparable across studies. Statistical evaluation, although more time-intensive due to the need for high-quality ground truth annotations and carefully designed test datasets, enables the use of clearly defined performance metrics, confidence intervals, and benchmarks, and facilitates rigorous comparison of different models and training strategies. At the same time, visual assessment remains indispensable for understanding the morphology of false predictions, identifying systematic failure modes, and guiding dataset refinement. In practice, visual and statistical evaluation provide complementary information, and a combination of both yields the most comprehensive understanding of overall performance and specific sources of error.

Reliable statistical performance evaluation requires more than just reporting a single metric on a convenient test set. It depends critically on the availability of an independent and representative ("real-world") test dataset, created with appropriate data splitting to avoid leakage, and with sufficient size and variability to capture the intended use case. Sample size considerations at the case level, together with confidence intervals and resampling methods such as bootstrapping, help to avoid overinterpretation of unstable point estimates, particularly in small or unbalanced test sets. Furthermore, thoughtful metric selection tailored to the pathology task, along with the use of multiple complementary metrics, is necessary to ensure that the reported performance truly reflects the clinically relevant aspects of the algorithm's behavior. When multiple models or configurations are compared, paired analyses, resampling-based confidence intervals for metric differences, and careful handling of multiple comparisons are essential to distinguish meaningful improvements from random variation or overfitting to a specific dataset. Finally, given the stochastic nature of DL training, repeated runs with different random seeds are crucial to assess model stability. When feasible, combining these multiple runs with case-level resampling quantifies uncertainty from both training stochasticity and finite test sample size

By outlining these considerations, we aim to support the development of good scientific practice in veterinary DL-AIA and to lay the groundwork for ongoing discussion within the veterinary pathology community on best practices for evaluating deep learning models.

# Acknowledgements , Conflict of interest, Funding

**Acknowledgement**

The authors acknowledge the use of ChatGPT (OpenAI, GPT-4) for assistance with proofreading and improving the clarity of the manuscript. The authors take full responsibility for the content, accuracy, and interpretation of the work presented in this publication.

**Declaration of Conflicting Interests**

The author(s) declared no potential conflicts of interest with respect to the research, authorship, and/or publication of this article.

**Funding**

The author(s) disclosed receipt of the following financial support for the research, authorship, and/or publication of this article: CAB acknowledges support from the Austrian Research Fund (FWF, project number: I 6555). JA acknowledges support by the Bavarian State Ministry for Science and the Arts (project FOKUS-TML). MA acknowledges support by the Deutsche Forschungsgemeinschaft (DFG, project number: 520330054).

43. Mehrvar S, Maisonave K, Buck W, Guffroy M, Bawa B, Himmel L. Immunohistochemistry-Free Enhanced Histopathology of the Rat Spleen Using Deep Learning. *Toxicol Pathol*. 2025;53**:** 83-94. 10.1177/01926233241303907
44. Meuten DJ, Moore FM, Donovan TA, et al. International Guidelines for Veterinary Tumor Pathology: A Call to Action. *Vet Pathol*. 2021;58**:** 766-794. 10.1177/03009858211013712
45. Naylor SW, McInnes EF, Alibhai J, Burgess S, Baily J. Development of a Deep Learning Tool to Support the Assessment of Thyroid Follicular Cell Hypertrophy in the Rat. *Toxicol Pathol*. 2025;53**:** 240-250. 10.1177/01926233241309328
46. Pacholec C, Flatland B, Xie H, Zimmerman K. Harnessing artificial intelligence for enhanced veterinary diagnostics: A look to quality assurance, Part I Model development. *Vet Clin Pathol*. 2024. 10.1111/vcp.13401
47. Palazzi X, Barale-Thomas E, Bawa B, et al. Results of the European Society of Toxicologic Pathology Survey on the Use of Artificial Intelligence in Toxicologic Pathology. *Toxicol Pathol*. 2023;51**:** 216-224. 10.1177/01926233231182115
48. Pischon H, Mason D, Lawrenz B, et al. Artificial Intelligence in Toxicologic Pathology: Quantitative Evaluation of Compound-Induced Hepatocellular Hypertrophy in Rats. *Toxicol Pathol*. 2021;49**:** 928-937. 10.1177/0192623320983244
49. Puget C, Ganz J, Ostermaier J, et al. Artificial intelligence can be trained to predict c-KIT-11 mutational status of canine mast cell tumors from hematoxylin and eosin-stained histological slides. *Vet Pathol*. 2025;62**:** 152-160. 10.1177/03009858241286806
50. Ramot Y, Deshpande A, Morello V, Michieli P, Shlomov T, Nyska A. Microscope-Based Automated Quantification of Liver Fibrosis in Mice Using a Deep Learning Algorithm. *Toxicol Pathol*. 2021;49**:** 1126-1133. 10.1177/01926233211003866
51. Ramot Y, Zandani G, Madar Z, Deshmukh S, Nyska A. Utilization of a Deep Learning Algorithm for Microscope-Based Fatty Vacuole Quantification in a Fatty Liver Model in Mice. *Toxicol Pathol*. 2020;48**:** 702-707. 10.1177/0192623320926478
52. Rosbach E, Ammeling J, Krügel S, et al.: "When Two Wrongs Don't Make a Right"-Examining Confirmation Bias and the Role of Time Pressure During Human-AI Collaboration in Computational Pathology. *In*: Proceedings of the 2025 CHI Conference on Human Factors in Computing Systems, pp. 1-18. 2025
53. Rudmann D, Albretsen J, Doolan C, et al. Using Deep Learning Artificial Intelligence Algorithms to Verify N-Nitroso-N-Methylurea and Urethane Positive Control Proliferative Changes in Tg-RasH2 Mouse Carcinogenicity Studies. *Toxicol Pathol*. 2021;49**:** 938-949. 10.1177/0192623320973986
54. Smith MA, Westerling-Bui T, Wilcox A, Schwartz J. Screening For Bone Marrow Cellularity Changes in Cynomolgus Macaques in Toxicology Safety Studies Using Artificial Intelligence Models. *Toxicol Pathol*. 2021;49**:** 905-911. 10.1177/0192623320981560
55. Stacke K, Eilertsen G, Unger J, Lundstrom C. Measuring Domain Shift for Deep Learning in Histopathology. *IEEE J Biomed Health Inform*. 2021;25**:** 325-336. 10.1109/jbhi.2020.3032060
56. Sullivan GM, Feinn R. Using Effect Size-or Why the P Value Is Not Enough. *J Grad Med Educ*. 2012;4**:** 279-282. 10.4300/jgme-d-12-00156.1
57. Tafavvoghi M, Bongo LA, Shvetsov N, Busund LR, Møllersen K. Publicly available datasets of breast histopathology H&E whole-slide images: A scoping review. *J Pathol Inform*. 2024;15**:** 100363. 10.1016/j.jpi.2024.100363
58. Tokarz DA, Steinbach TJ, Lokhande A, et al. Using Artificial Intelligence to Detect, Classify, and Objectively Score Severity of Rodent Cardiomyopathy. *Toxicol Pathol*. 2021;49**:** 888-896. 10.1177/0192623320972614
59. Vuorimaa M, Kareinen I, Toivanen P, Karlsson S, Ruohonen S. Deep Learning-Based Segmentation of Morphologically Distinct Rat Hippocampal Reactive Astrocytes After Trimethyltin Exposure. *Toxicol Pathol*. 2022;50**:** 754-762. 10.1177/01926233221124497
60. Wasserstein RL, Schirm AL, Lazar NA: Moving to a world beyond "p< 0.05", pp. 1-19. Taylor & Francis, 2019

# Supplemental Materials – Best Practice Recommendations

**Supplementary Table S1**. Best practice recommendation for rigorous performance evaluation of DL-AIA in veterinary pathology.

| Aspect | Best practice recommendation | Rationale / notes |
|---|---|---|
| **Overall evaluation approach** | | |
| Overall evaluation strategy | Combine statistical evaluation with targeted visual assessment of algorithmic predictions. | Statistical methods provide transparent, reproducible, and comparable performance estimates; visual assessment is indispensable for understanding error modes and guiding dataset refinement. |
| Primary evaluation method | Use statistical performance evaluation with clearly defined metrics wherever feasible; use exclusive visual evaluation only with a strong rationale why metrics are not applicable. | Statistical evaluation scales to large datasets, supports benchmarking and model comparison, and aligns with veterinary and human pathology guidelines (e.g., veterinary AI reporting guidelines). |
| Role of visual assessment | Use visual assessment to explore sources of error, prioritize failure modes, and refine training and test datasets (e.g., identifying difficult regions, artifacts, domain shifts). | Visual inspection can reveal systematic error patterns (e.g., false positives in necrotic/inflamed areas) that metrics alone do not explain. |
| Transparency and communication | Report performance using explicit metrics (e.g. F1-score, Dice, AUC) with clear definitions and thresholds; avoid vague terms such as "high" or "sufficient" performance. | Quantitative, well-defined metrics are interpretable, reproducible, and comparable across studies; qualitative labels are subjective and non-transferable. |

| Interpretability and benchmarks | Whenever possible, define benchmarks (e.g. performance of pathologists or prior state-of-the-art models) for the same images and metrics. | Benchmarks are necessary to interpret whether a given metric (e.g. F1 = 0.7) is acceptable for the specific task, dataset, and ground truth. |
|---|---|---|
| **Test dataset design and independence** | | |
| Test set independence and data leakage | Use a hold-out test set that is strictly independent from training and validation; reserve all images from a given patient/animal to a single subset to avoid data leakage. | Avoids data leakage (e.g. tiles from same slide in train and test) and thereby prevents overly optimistic performance estimates driven by spurious correlations. |
| Representativeness of test data | Design test data to be representative of the intended real-world use case, including variability in tissue, disease subtypes, slide preparation, staining, WSI scanners, artifacts, and case difficulty, if applicable. Avoid methods like active learning and synthetic data to construct the test images sets. | Performance on "perfect" regions or homogeneous samples may not translate to routine diagnostic workflows; representativeness is essential for external validity. |
| Test set size and precision | Consider simple sample size / precision calculations for key metrics at the level of independent units, aiming for sufficiently narrow confidence intervals around sensitivity, specificity, etc. | Small test sets can yield unstable estimates where wide confidence intervals make strong conclusions impossible; case-level planning reduces overinterpretation. |
| **Ground truth quality and imperfect reference standards** | | |
| Ground truth quality | Use rigorous annotation strategies for test sets (e.g. multiple experts with blinded majority voting, structured decision guides, computer-assisted candidate identification). | Reduces label noise and systematic bias in ground truth, which otherwise caps achievable model performance and can misrepresent algorithm quality. |

| Recognizing imperfect reference standards | Explicitly acknowledge that pathologist annotations reflect aleatoric and epistemic uncertainty, and interpret model performance relative to this imperfect reference. | Makes clear that discrepancies may reflect ambiguous cases or observer variability, not only algorithmic failure. |
|---|---|---|
| Algorithmic labels and semi-automatic annotation | When using algorithmically generated labels to accelerate annotation, ensure expert validation and correction of labels, especially in the test set. | Unverified algorithmic labels can propagate errors, degrade ground truth quality, and bias performance estimates. |
| **Metric selection and reporting** | | |
| Choice of metrics | Select metrics according to the pattern recognition task, e.g. classification, segmentation, or object detection. | Different metrics emphasize different aspects of performance; inappropriate metrics can misrepresent clinically relevant behavior. Use decision guides (e.g. Metrics reloaded) where available. |
| Aligning metrics with pathology task | Ensure metrics reflect the clinical task: e.g., for cell counting, consider both segmentation quality and object detection; for grading tasks, consider clinically meaningful thresholds and misclassification patterns. | Good segmentation metrics do not guarantee good counting or clinical decision-making, and vice versa. |
| **Uncertainty quantification** | | |
| Uncertainty quantification | Accompany key performance metrics with confidence intervals, preferably via resampling methods such as bootstrapping at the case level. | Confidence intervals convey the precision and robustness of performance estimates, especially critical for small or unbalanced test sets. |

| Bootstrapping strategy | Perform bootstrapping by resampling independent units (patients, animals, WSIs) with replacement and recomputing aggregate metrics for each resample; use the resulting empirical distributions to derive CIs. | This approach applies uniformly whether metrics are viewed as "case-level" or "aggregate" and correctly reflects uncertainty due to finite test sample size. |
|---|---|---|
| **Statistical comparison of models** | | |
| Model comparison - paired structure | When comparing multiple models on the same test cases, treat predictions as paired and use paired methods (e.g. McNemar's test for classification, paired t-test or Wilcoxon for continuous outcomes, or paired bootstrap of Δmetrics). | Ignoring the paired nature (same cases for both models) discards information and may bias conclusions; paired tests and paired resampling exploit the shared structure. |
| Model comparison - resampling Δmetrics | For comparing models by aggregate metrics (e.g. F1, Dice, AUC), use case-level bootstrap resampling with shared resamples for both models and compute Δmetric for each resample to obtain a confidence interval for the difference. | Provides a direct, interpretable estimate of whether performance differences exceeded what could be expected from random variation in the finite test set. |
| Multiple comparisons | If many models / configurations are compared on the same test set, acknowledge the multiple comparison problem and apply corrections (Bonferroni, Holm, FDR) or clearly label the analysis as exploratory. | Testing many models increases the chance of false-positive "winners"; correction or cautious interpretation reduces spurious claims of superiority. |

| Focus on effect sizes and confidence intervals | Emphasize effect sizes and confidence intervals for differences in performance (e.g. ΔF1) over p-values alone, especially when sample sizes are large or many comparisons are performed. | P-values can be very small for trivial differences in large datasets; effect sizes and confidence intervals better capture clinical relevance and robustness. |
|---|---|---|
| Clinical / biological relevance | Interpret performance differences with regard to clinical or biological significance (e.g. impact on diagnostic accuracy, workload, or outcomes), not only statistical significance. | Avoids overinterpretation of negligible yet statistically "significant" improvements. |
| **Multiple training runs and model stability** | | |
| Multiple training runs | For each candidate configuration, if feasible, perform multiple independent training runs with different random seeds and summarize performance across runs (mean, SD, range). | Training is stochastic; single-run results may be overly optimistic or pessimistic. Multiple runs assess model stability. |
| Interpreting few runs (e.g. 3-5) | When only a small number of runs are feasible, treat them primarily as a robustness / sensitivity check, not as a strong basis for formal hypothesis testing between configurations. | With very few runs, paired tests have low power and unstable p-values; descriptive summaries are more honest. |
| Combining runs with resampling | When possible, combine multiple runs with case-level bootstrapping (K runs × B resamples) to estimate uncertainty that reflects both training variability and sampling variability. | Produces more realistic confidence intervals and Δmetrics distributions than single-run, single-split analyses. |
| **Subgroup analysis** | | |

| Hidden stratification and subgroup analysis | Perform stratified evaluation by relevant subgroups (e.g. scanner, lab, tissue subtype, artifact presence) to detect hidden performance deficits. | Aggregate metrics can mask failures in clinically important subgroups; stratification uncovers such issues. |
|---|---|---|
| **Data sharing and reproducibility** | | |
| Open datasets and reproducibility | Whenever possible, make test datasets (and ideally full datasets) publicly available, with clear documentation of ground truth methods and inclusion criteria. | Enables external validation, fair model comparison across studies, and facilitates methodological research. |
| Risk of overfitting to public benchmarks | When a small public datasets has been repeatedly used by numerous studies, acknowledge the risk that incremental reported performance gains may partly reflect hyperparameter overfitting to the benchmark rather than true methodological advance. | Encourages cautious claims and motivates the creation and sharing of additional, diverse datasets. |